# MatGD: Materials Graph Digitizer

Jaewoong Lee[1,†], Wonseok Lee[1,†], Jihan Kim[1]

[1]Department of Chemical and Biomolecular Engineering, Korea Advanced Institute of Science and Technology, Daejeon, Republic of Korea

* Correspondence to: Jihan Kim (jihankim@kaist.ac.kr)

[†]*These authors contributed equally to this work.*

# Abstract

We have developed MatGD (Material Graph Digitizer), which is a tool for digitizing a data line from scientific graphs. The algorithm behind the tool consists of four steps: (1) identifying graphs within subfigures, (2) separating axes and data sections, (3) discerning the data lines by eliminating irrelevant graph objects and matching with the legend, and (4) data extraction and saving. From the 62,534 papers in the areas of batteries, catalysis, and MOFs, 501,045 figures were mined. Remarkably, our tool showcased performance with over 99% accuracy in legend marker and text detection. Moreover, its capability for data line separation stood at 66%, which is much higher compared to other existing figure mining tools. We believe that this tool will be integral to collecting both past and future data from publications, and these data can be used to train various machine learning models that can enhance material predictions and new materials discovery.

# Introduction

Materials sciences, chemistry, and other related fields are undergoing a rapid transformation due to the explosion of digital data generated via experiments and computational simulations. As such, with the advancements of machine learning and deep learning algorithms, it remains important to utilize data-driven techniques in the field. Specifically, these advanced algorithms can be deployed on the large materials datasets to accurately predict material properties, optimize synthesis routes, and even discover new materials.[1–5]

However, majority of the data currently available originates from computational simulations[3,6–9], which can sometimes offer misleading insights that do not align with experimental findings. Due to this concern, there is a growing interest in acquiring a large set of experimental data to train the machine learning algorithms to enhance the accuracy of the predictions. Despite its importance, collecting experimental data is fraught with challenges; it is often expensive, time-consuming, and labor-intensive.[10–12] This bottleneck has led researchers to turn their attention to the abundant amount of textual data found in scientific publications. Leveraging this textual data via sophisticated mining techniques not only supplements computational models but also enriches our understanding of real-world material behavior.

To this end, several researchers have demonstrated innovative approaches in this space. Notable contributions in this area include the work by Park et al.[13], who developed a targeted text-mining algorithm specifically for metal–organic frameworks (MOFs). Their algorithm successfully identified surface area and pore volume values with impressive accuracies of 73.2% and 85.1%, respectively, for a test set of MOF html files. Furthermore, Kononova et al.[14] utilized text-mining approaches in conjunction with natural language processing to create a dataset of 19,488 solid state synthesis entries, thereby providing an invaluable resource for the development of synthesis routes in inorganic materials. Likewise, Nandy et al.[15] combined natural language processing and machine learning to predict the stability properties of MOFs, utilizing a comprehensive dataset gathered from approximately 4,000 manuscripts. Park et al. in 2022 applied machine learning in tandem with text-mining to achieve an average F1 score of 90.3% in predicting MOF synthesis parameters such as solvents, temperature, and metal precursors. Most recently, Glasby et al. generated the DigiMOF[16] database using a chemistry-aware natural language processing tool, offering an open-source repository of MOF synthetic data gleaned from over 43,000 journal articles.

While these strides in text mining and resulting accumulation of text data have been significant, data that is found from images and figures have been ignored for the most part. This is a substantial oversight, given that figures often encapsulate the critical results of a study, and this data is not often found in the text sections in a given publication. There are only a few cases where figure mining has been attempted.[16–18] In particular, Nandy et al.[15] digitized TGA data for approximately three thousand metal-organic framework papers. While their success is noteworthy, to expand this to a broader scope in materials science, there exist a pressing need to develop tools that can facilitate obtaining data found in figures.

To remedy this issue, a few groups have developed tools that can mine data from figures. However, currently existing figure mining tools[19–23] face limitations in dealing with the specific demands of data mining tasks. For instance, they may perform well on synthetic datasets but struggle with real-world data from published materials science research. Additionally, meaningful utilization of graph data often requires understanding the nuances of data lines conveyed through legends or text within the graph, which poses another challenge.

In response to these issues, we introduce MatGD, which is a figure mining tool developed to handle quantitative data from scientific publications. During its development, we utilized real-world data from published materials research papers to ensure its effectiveness on real data, and not just synthetically generated ones. MatGD also incorporates a new algorithm for information mapping, which aids in understanding and applying the data within graphs. By addressing these challenges, we believe that MatGD will bridge the gap in figure mining within the field of materials science, offering a new avenue for materials discovery and development.

## Methods

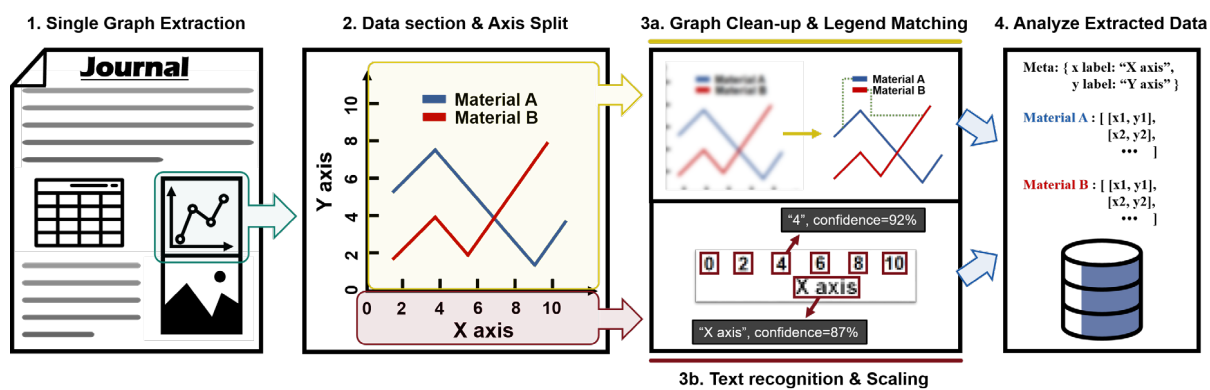

**Figure 1. The overall process of mining process**

The overall workflow of our program consists of four steps (see Figure 1). Our program takes in a figure as input (which can be downloaded from links in the paper's XML/HTML or extracted from a PDF of a paper) and can handle many different formats including JPG and PNG. In scientific papers, figures commonly consist of multiple subfigures (e.g. Figure 1a, 1b, 1c, 1d) and these subfigures can be categorized into graphs (i.e. figures with numerical x and y axis) and non-graphs. Throughout the work, the YOLO[24] (You Only Look Once) model is primarily employed for detection tasks that serve as precursors for subsequent classification, extraction, or removal of objects. For objects that pose challenges for YOLO detection due to their specific characteristics, we utilize rule-based algorithms.

Initially, figures are divided into individual subfigures, which are then classified as either graph or non-graph. After filtering out non-graphs, the remaining graphs undergo further segmentation into their data section area, x-axis section, and y-axis section. Within the data section area, any objects other than the main data points/lines are deleted to isolate just the data. After this, the remaining data lines are clustered together based on their color, and each of these data lines is mapped with its corresponding legends and texts. After extracting numerical values from the axes, the obtained data lines are rescaled to the values of the original image. In the next subsections, we detail each of the four main steps that consist of our program.

*Single graph extraction*

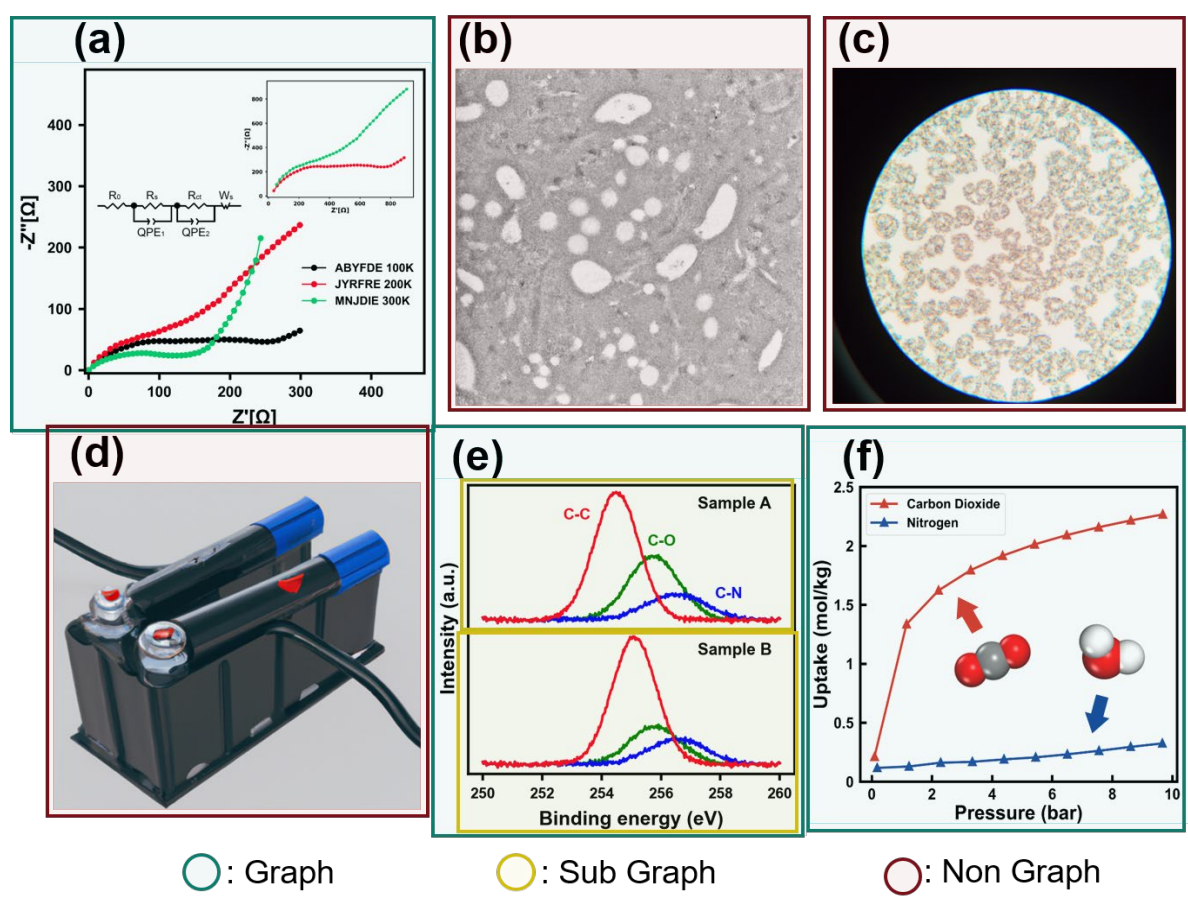

**Figure 2. Example of separating graphs and non-graphs in figure. Parts (b), (c) and (d) are generated using OpenAI's DALL-E 2.[25] Parts (a), (e), and (f) are created with artificial data and are not sourced from any publication.**

For our object detection tasks, we chose YOLO due to its many qualities that are particularly advantageous for mining data from a large number of scientific papers. Unlike other popular object detection algorithms such as Faster R-CNN[26], which employs a two-stage process for object localization and classification, or SSD[27], and RetinaNet[28], which balance speed and accuracy but can be complex to implement, YOLO performs both tasks simultaneously. This results in significant speed advantages, making it highly suitable for handling large datasets. Additionally, the YOLO model is relatively straightforward to set up and integrate into our workflow, simplifying the overall process. We specifically employed the YOLOv8x model, which boasts 68.2 million parameters and is the latest iteration known for top-tier performance. By its 8th version, YOLO's accuracy has not only become comparable to that of its CNN-based counterparts like Faster R-CNN, SSD, and RetinaNet but has also outperformed them depending on the characteristics of the objects being detected.[29,30]

In the application of the YOLO model, the image pixel size was set to 640. This parameter was optimized based on the computational constraints imposed by our available GPU resources, while also satisfying the requisite accuracy criteria. (It should be noted that this pixel size corresponds with the default configuration for the YOLO model.) Subsequent to these settings, the model was trained on a manually-curated dataset to ensure the quality and relevance of the data employed. These datasets comprise a cumulative total of 2,047 figures, which are further segmented into 5,974 subfigures. The aforementioned figures collectively represent the training, validation, and testing sets, partitioned in a ratio of 4:1:1, respectively. The YOLO model classifies each subfigure as either 'graph' or 'non-graph' and simultaneously returns the bounding boxes for the appropriate regions as depicted in Figure 2. In the figure, we introduce an additional classification referred to as a 'Sub Graph.' A Sub Graph is defined as a distinct graph within a single overarching graph that shares either the x-axis or y-axis with the parent graph. We have incorporated this classification to facilitate subsequent analyses.

***Splitting data section & axis***

Subfigures classified as 'graphs' undergo a detailed analytical process to isolate their constitutive elements: the abscissa (x-axis), the ordinate (y-axis), and the data section of the graph. This segmentation is executed using a rule-based algorithm developed using the OpenCV[31] library in Python. Our algorithm is designed to perform line detection within the graphical representation, employing specific criteria concerning the line length and its proximity to the extremal boundaries of the figure. Lines that both exhibit the greatest length and are nearest to the periphery are predominantly identified as the axes, thereby demarcating their respective regions within the graph.

It is noteworthy to specify that while the YOLO model was successfully employed in other stages of our study, it yielded unsatisfactory outcomes when applied specifically to the segmentation of the x-axis and y-axis in scientific graphs. This limitation was particularly evident in YOLO's difficulty with recognizing and detecting lines or objects with unclear boundaries, a feature commonly found in the axes of scientific graphs. For the same reason, we later used another algorithm, the Density-Based Spatial Clustering of Applications with Noise (DBSCAN) instead of YOLO to separate data lines in the graph's data section.

Following the axis identification, the remaining unmarked region is designated as the data section. This region is subjected to further processing steps, including data artifact removal and legend association, to facilitate subsequent analysis. Textual content within the axes is extracted using text recognition techniques, providing critical context such as labels, units, and scale for a more comprehensive understanding of the graph.

***Graph clean-up & legend matching***

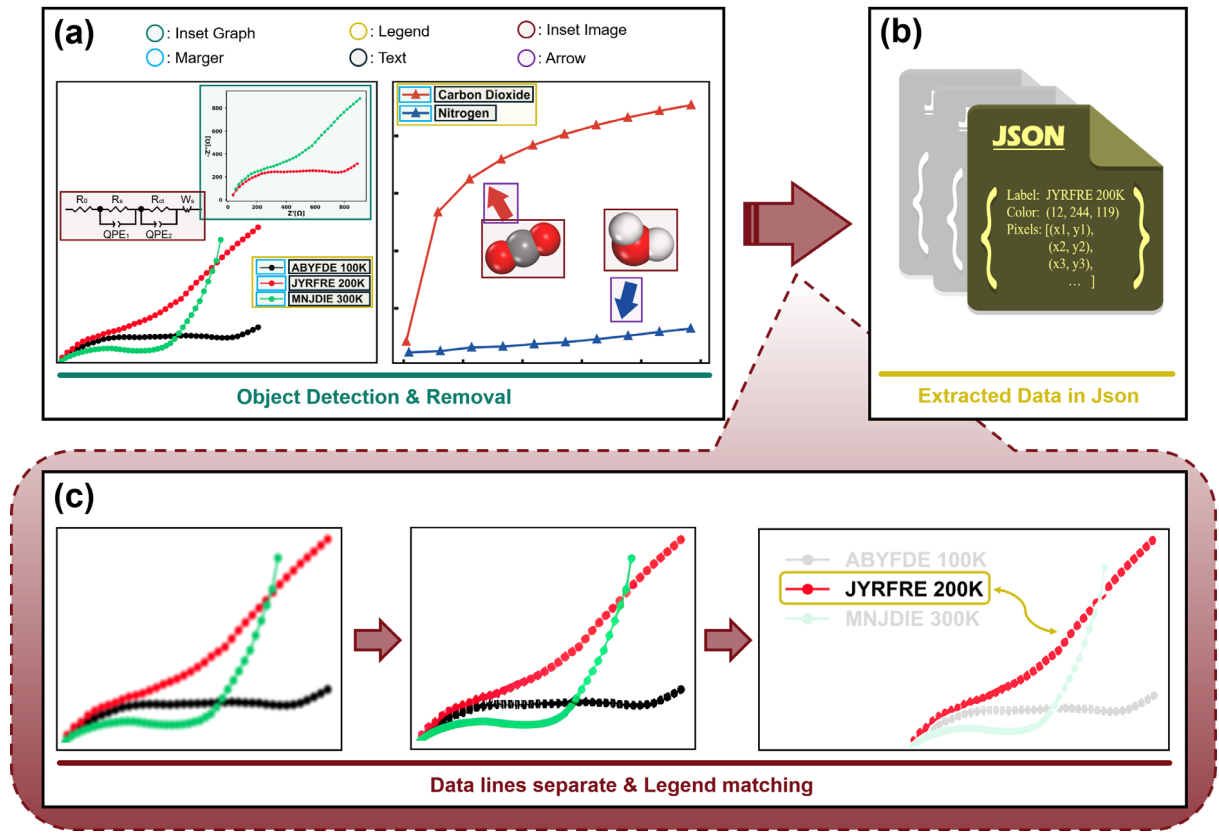

**Figure 3. Illustration of graph clean-up & legend matching**

Similar to the single graph extraction process, we employed the YOLOv8x model (68.2M, 640 pixel size) to detect objects in the graph's data section. Specifically, the YOLO model was trained to detect six object classes (i.e. 'text', 'marker', 'legend', 'arrow', 'inset image', and 'inset graph'). A total of 4,579 single-graph images were manually labeled and partitioned into training, validation, and testing sets for the purpose of the classification procedure. A diverse dataset of figures was assembled to ensure that each of the six predefined object classes was represented by more than 650 instances. It should be noted that data points and lines are not included within the bounding boxes and are thus excluded from the YOLO training model(refer to Figure 3(a) for an example that illustrates the parsing of these six object classes).

In addition to the six object classes defined earlier, we devised a rule-based script to detect ticks and grids. Ticks are small graphical elements used to mark intervals on the axes, and grids are lines that guide the eye across the data plots. Due to their small size and line-like characteristics, these elements pose challenges for detection, which are inadequately addressed by the YOLO model. Our rule-based algorithm is designed to surmount these limitations: it identifies ticks by detecting small rectangles at regular intervals along the boundaries of the image and locates grids by recognizing lines that are closely aligned with the horizontal and vertical axes at uniform intervals.

After removing graph objects, ticks, grids, we were able to isolate the data lines. As illustrated in Figure 3(c), we implemented additional techniques to facilitate subsequent data processing and further improve detection accuracy. First, we converted any background areas with RGB values above a specific threshold to pure white to enhance contrast. Secondly, we used the ESRGAN[32] model to obtain high-resolution images. To match the markers of legend with the data lines, it is essential to first separate the data lines. DBSCAN algorithm was used to separate data lines based on their RGB color values. The parameters for DBSCAN were calibrated to account for the distinct features inherent to each image (e.g. image size, ratio of colored pixels). To extract information from the data line, we applied the algorithm described below. When the number of data lines matched the number of legend markers, we employed cosine similarity to map the RGB values of the markers to the RGB values of the corresponding data lines. For text in graph, we introduced a simple machine learning model that takes the minimum distance between the text and data lines and RGB value of those as inputs and predicts the mapping relationship between them as the output. We used over 1000 text-data relationship pairs as training data. Matched text-data is finally cleaned up in JSON format, as shown in Figure 3(b).

### Text recognition & Scaling

We incorporated the RobustScanner[33] model for text recognition. It was used to recognize the text within the data section, labels of axes and numerical values along the axes.

Numerical characters on the axes were extracted using the text recognition model. These values were then rescaled to match their true values in the original image. During this process, we obtained the pixel positions of each numerical character. The accuracy of the extraction was validated by checking whether the numerical values and their corresponding pixel positions conformed to an arithmetic sequence. If the scale of the axis was arbitrary, data were extracted as normalized values.

## Results

### *Database*

We amassed a collection of over 500,000 figures extracted from 62,534 articles sourced from Elsevier journals. These papers span research fields including Metal-Organic Frameworks (MOFs), batteries, and catalysts. To categorize the subfigures, we employed the previously mentioned YOLO model, which discriminates between graph and non-graph. A detailed breakdown of this classification is provided in the subsequent table.

**Table 1 The material database used for the development of the tool.**

| Type of Material | Total Number of | | | | |
|---|---|---|---|---|---|
| | Papers | Figures | Sub-Figures | Graphs | Images |
| Battery | 23,099 | 159,859 | 579,344 | 314,846 | 264,498 |
| MOF | 19,380 | 161,682 | 552,776 | 262,265 | 290,511 |
| Catalyst | 20,055 | 178,442 | 627,491 | 343,657 | 283,834 |
| All | 62,534 | 501,045 | 1,759,611 | 920,768 | 838,843 |

### *YOLO Models*

Table 2 presents the accuracy results for the two YOLO models used in single graph extraction and object detection. The number in parentheses next to each class represents the count of instances used for training. Mean average precision (mAP) is a metric that provides a single number to evaluate the overall performance of an object detection model across all classes. The mAP value is calculated using the following equation by dividing the average precision, obtained from the precision-recall curve, by the number of classes.

$$\text{mAP} = \frac{1}{n}\sum_{k=1}^{k=n} AP_k \quad (1)$$

In equation (1), $AP_k$ denotes the average precision of class $k$, and $n$ represents the total number of classes. The subscripted number following mAP represents the Intersection over Union (IoU) threshold. It is used to determine whether a prediction in object detection is considered correct when the overlapping area between the prediction and ground truth exceeds that threshold. '50:90' denotes the average value obtained by varying the threshold from 50 to 90.

As mentioned in the Method section, during the development of the YOLO model for single graph extraction, we manually labeled 5,974 subfigures from a total of 2,047 figures. This dataset was partitioned into training, validation, and test sets at a ratio of 4:1:1 for the verification process. Similarly, for the graph object detection, we labeled instances obtained from 4,579 single graphs and conducted testing after splitting them in a 4:1:1 ratio.

In the single graph extraction process, YOLO models demonstrated excellent accuracy in classifying graph, non-graph, and subgraph categories. In object detection process, we observed differences in mAP values between objects. Accuracy for arrow and marker was lower than other objects, primarily because arrow instances were less prevalent in the training data due to the inherent nature of graphs (Figure S1). This data imbalance could potentially have a negative impact on the training process. In the case of markers, they often shared a similar shape with scatter points in data lines, which is also considered a reason of errors

**Table 2. The object detection accuracy of the YOLO model.**

| Classification | Criteria | | |
| --- | --- | --- | --- |
| | $mAP_{50:95}$ | $mAP_{50}$ | Box |
| Graphs | 0.955 | 0.976 | 0.968 |
| Sub-Graphs | 0.920 | 0.965 | 0.937 |
| Non-Graphs | 0.827 | 0.906 | 0.893 |
| All | 0.901 | 0.949 | 0.932 |

| Classification | Criteria | | |
| --- | --- | --- | --- |
| | $mAP_{50:95}$ | $mAP_{50}$ | Box |
| Text | 0.810 | 0.987 | 0.966 |
| Marker | 0.727 | 0.994 | 0.991 |
| Legend | 0.907 | 0.992 | 0.970 |
| Arrow | 0.613 | 0.921 | 0.945 |
| Inset Graph | 0.981 | 0.995 | 0.985 |
| Inset Image | 0.907 | 0.964 | 0.947 |
| All | 0.824 | 0.976 | 0.967 |

*Comparison to Other Tool*

We compared our solution with the LineEX tool, which is currently the most up-to-date and accurate tool to the best of our knowledge. Our comparison evaluated accuracy in three tasks: (1) legend marker detection, (2) legend text detection, and (3) data line separation. In the case of legend marker detection and legend text detection, a detection box was considered correct if it had an Intersection over Union (IoU) score of 0.7 or higher with the manually labeled ground-truth box and was correctly classified. For data line separation, the result was considered accurate only when the number of data lines separated by DBSCAN matched the number of legends in the original image, and the distance between the RGB value vectors of the detected lines and the legends was sufficiently close.

We utilized two types of graph data in comparison. One type is the LineEX graph[20], generated using the Python Matplotlib library, and the other is the material paper graph, extracted from scientific journals. Table 3 and Table 4 present the accuracy values for LineEX graphs and material paper graphs, respectively. In both graph types, we utilized 100 single graph images as the test set. For LineEX graphs, both our tool and LineEX exhibited commendable performance. However, in the case of materials paper graphs, a stark performance difference between the two tools was evident. Our tool showcased superior performance in all three tasks, legend marker detection, text detection, and data line separation.

**Table 3 Results of the accuracy test conducted on the LineEX graph dataset**

| Tool | Accuracy (for each of the 423 instances) | | |
|---|---|---|---|
| | Legend marker detection | Legend text detection | Data line separation |
| MatGD(ours) | 89.7% | 99.5% | 79.2% |
| LineEX | 83.7% | 85.8% | 61% |

**Table 4 5 Results of the accuracy test conducted on the material paper graph dataset**

| Tool | Accuracy (for each of the 340 instances) | | |
|---|---|---|---|
| | Legend marker detection | Legend text detection | Data line separation |
| MatGD(ours) | 99.4% | 99.7% | 66.1% |
| LineEX | 44.9% | 55.4% | 10.6% |

*Mining Examples*

Next, we provide examples of data mined for three materials categories used in our database: Metal Organic Frameworks (MOFs), batteries, and catalysts. Figure 5 illustrates these examples. For MOFs, we plotted the $N_2$ adsorption isotherm at 77 K. $N_2$ adsorption is a commonly used method for measuring the pore volume of MOFs. Usually, these graphs contain two lines, representing adsorption and desorption, making it convenient for mining with our tool. Additionally, extracting the $N_2$ isotherm graphs required more than just the information within the graphs themselves. Therefore, we utilized the captions that were downloaded alongside the figures when we obtained them and applied chemical knowledge. Initially, we filtered graphs by identifying captions containing '$N_2$' or 'Nitrogen'. Subsequently, we verified axis labels corresponding to pressure and pore volume. Given that $N_2$ adsorption measurements typically cover a range up to 1 bar, we established criteria based on the x-axis range.

In battery research papers, numerous graphs depict the performance and properties of materials. Among these, we plotted the distribution of capacity, as shown in Figure 5(b). This is analogous to a study that mined the distribution of various properties of battery materials using the text mining tool ChemDataExtractor[34]. The labels of axes were obtained through text recognition. Unlike captions, the text strings obtained through text recognition may contain typographical errors. Therefore, we introduced the Levenshtein algorithm for mining graphs where the x-label represented cycle number and the y-label represented capacity.

In the case of catalysts, the type of graphs representing properties varies for each catalytic reaction. We selected the hydrogen evolution reaction (HER) as a representative example. In Figure 5(c), we presented the Gibbs free energy for catalysts used in the HER. Similar to the previous graphs, we filtered the captions of graphs that contained content related to the hydrogen evolution reaction. Subsequently, we obtained graphs where the x-label represented the reaction pathway or reaction coordination. Overall, aggregation of these data across different fields of study showcase a powerful utility of our tool that can expedite collecting thousands of data from figures and help facilitate understanding trends as well as unearthing structure-property relationships for many different materials.

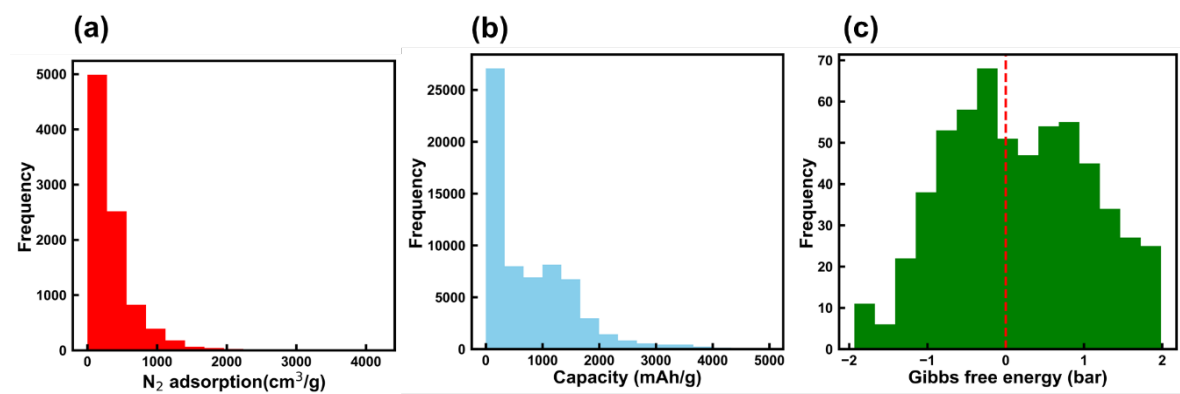

**Figure 4.** Mining examples correspond to MOF (a), battery (b), and catalyst (c).

***Advantages and Limitations***

MatGD generally performs well across various data line characteristics, and it excels when data lines are represented by a limited number of colors or when there is a distinct difference in the RGB values of these colors. Notable examples of such graphs in the field of materials science include X-ray diffraction (XRD), adsorption isotherm, and Thermalgravimetric Analysis (TGA). However, it may encounter challenges and exhibit suboptimal performance in specific instances. MatGD's performance diminishes when confronted with data lines represented by similar colors, such as shades of gray or black, or when there is significant overlap between data lines, as often found in X-ray photoelectron spectroscopy (XPS).

**Conclusion**

In this study, we introduced MatGD, a tool designed to mine data from line graphs, particularly in materials research papers. Integrating the renowned YOLO framework with additional sophisticated techniques, we've tailored a potent mining algorithm that stands out for its efficacy. Our tool performs effectively on real-world data extracted from published materials research papers, making it a valuable asset for leveraging mined data. We anticipate that the utilization of our tool will foster increased research activity in the field of figure mining within the domain of materials science.

# ACKNOWLEDGEMENT

This work was supported by the National Research Foundation of Korea (NRF) under Project Number 2021M3A7C208974513, 2019M3A7B4043807.

# ASSOCIATED CONTENT

**Supporting Information**. The following files are available free of charge.

# AUTHOR INFORMATION

**Corresponding Author**

**Jihan Kim** – *Department of Chemical and Biomolecular Engineering, Korea Advanced Institute of Science and Technology, Daejeon 34141, Republic of Korea;* orcid.org/0000-0002-3844-8789; Email: jihankim@kaist.ac.kr

**Authors**

**Jaewoong Lee** – *Department of Chemical and Biomolecular Engineering, Korea Advanced Institute of Science and Technology, Daejeon 34141, Republic of Korea;* orcid.org/0009-0002-8968-3292; Email: skyljw0714@kaist.ac.kr

**Wonseok Lee** – *Department of Chemical and Biomolecular Engineering, Korea Advanced Institute of Science and Technology, Daejeon 34141, Republic of Korea;* orcid.org/0000-0001-9624-3717; Email: ws1715@kaist.ac.kr

**Author Contributions**

J. L and W. L conceived the project, developed the program, and wrote the paper. J. K. directed the project.

**Notes**

The authors declare no competing financial interest.